\pdfoutput=1

\documentclass[11pt]{article}

\usepackage{EMNLP2022}

\usepackage{times}
\usepackage{latexsym}
\usepackage{booktabs}
\usepackage{subcaption}
\usepackage{mwe}
\usepackage{comment}
\usepackage{hyperref} 

\usepackage{amsmath,amssymb}

\DeclareMathOperator*{\argmin}{arg\,min}

\DeclareMathOperator{\EX}{\mathbb{E}}

\usepackage[T1]{fontenc}

\usepackage[utf8]{inputenc}

\usepackage{microtype}

\usepackage{inconsolata}

\usepackage{xcolor}

\title{Attack on Unfair ToS Clause Detection: A Case Study using Universal Adversarial Triggers}

\author{Shanshan Xu \and Irina Broda \and {\bf{Rashid Haddad}} \\ {\bf{Marco Negrini}} \and {\bf{Matthias Grabmair}} \\
School of Computation, Information, and Technology; Technical University of Munich, Germany\\
\texttt{\{firstname.lastname\}@tum.de} \\
    }

\begin{document}
\maketitle

\begin{abstract}
Recent work has demonstrated that natural language processing techniques can support consumer protection by automatically detecting unfair clauses in the Terms of Service (ToS) Agreement. This work demonstrates that transformer-based ToS analysis systems are vulnerable to adversarial attacks. We conduct experiments attacking an unfair-clause detector with universal adversarial triggers. Experiments show that a minor perturbation of the text can considerably reduce the detection performance. Moreover, to measure the detectability of the triggers, we conduct a detailed human evaluation study by collecting both answer accuracy and response time from the participants.  The results show that the naturalness of the triggers remains key to tricking readers.  	
\end{abstract}

\section{Introduction}
When using online platforms, users are asked to agree to the Terms of Service (ToS), which are often long and difficult to understand. According to \cite{obar2020biggest}, it would take a user around 45 minutes on average to read a ToS properly. Most users accept the terms without reading them, including clauses which would be deemed unfair under consumer protection standards. Software applications that warn consumers about unfair clauses can support consumers' rights, and have been the subject of prior work (e.g., \citealp{lippi2019claudette, ruggeri2022detecting}). At the same time, their existence forms an incentive for drafters of ToS to formulate clauses with potentially unfair effects that bypass automated screening. In turn, developers of control systems seek to make their detectors robust against such `adversarial attacks'. In this paper, we report on an experiment in discovering weaknesses of ToS analysis models.\\
Natural language processing (NLP) models for ToS analysis conduct binary classification of a given clause as fair/unfair. Previous studies have shown that state-of-the-art transformer-based classifiers are vulnerable to adversarial attacks \cite{belinkov2017synthetic}; even slight modifications to the input text (e.g., changing a few characters) can cause incorrect classifications \cite{ebrahimi-etal-2018-hotflip}. Numerous adversarial attack methods have been developed and demonstrate effective attack performance in various downstream NLP tasks such as sentiment analysis \cite{iyyer-etal-2018-adversarial}, question answering \cite{wang-etal-2020-t3}, machine-translation \cite{cheng-etal-2019-robust} etc. 
One such method is the attack via a \textit{universal adversarial trigger}, which is a sequence of tokens (words, sub-words, or characters) that can be injected into \textit{any} text input from a dataset to mislead the victim model to a target prediction (see Table \ref{tab:uat_example} for examples). These input-agnostic triggers, once generated, can be distributed to anyone,  and  do not need access to the victim model at the time of attack.\\
Adversarial attacks have, to the best of our knowledge, remained largely unaddressed in legal NLP. Our work extends the state of the art through the following contributions:
(1) We conduct experiments attacking ToS unfair clause detectors trained on the public CLAUDETTE dataset with universal adversarial triggers. Our results show that a minor perturbation of the text can reduce the detection performance of transformer based models significantly.
(2) We also use artifacts from the training data for universal trigger attacks. Our experiments demonstrate that such words can considerably reduce the victim model's accuracy, highlighting the potential threat of training data leakage.
(3) We conduct a human evaluation study to measure the detectability of the generated triggers. The results show that suppressing sub tokens can make generated triggers more difficult to detect.  \footnote{Our code is available at \url{https://github.com/TUMLegalTech/ToS_attack_nllp22}}



\begin{table*}[]
\begin{tabular}{|l|l|}
\hline
\textbf{ToS Clause   (\textcolor{red}{red} = trigger) }  &\textbf{Model Detection} \\ \hline
\begin{tabular}[c]{@{}l@{}}Pinterest isn't liable for damages that result from a \textbf{\textcolor{red}{may vote against}} non-material \\breach of any other applicable duty of care.\end{tabular} & Unfair $\rightarrow$ Fair\\ \hline
\begin{tabular}[c]{@{}l@{}}The English courts will have jurisdiction over any claim arising from \textbf{\textcolor{red}{may vote}} \\ \textbf{\textcolor{red}{against}}, or related to , any use of our services.\end{tabular} & Unfair $\rightarrow$ Fair\\ \hline
\end{tabular}
\caption{The universal adversarial trigger can be injected into \textit{any} input from a dataset to mislead the victim model. By inserting the displayed trigger can cause the trained unfair ToS detector to flip its correct unfair predictions to fair.}
\label{tab:uat_example}
\end{table*}

\section{Related Work}
\textbf{Adversarial Attacks in NLP}: Most adversarial attack methods in NLP are white-box, where the attacker has full access to the victim model (including architectures, parameters, and training data). Prevalent white-box attacks include HotFlip \cite{ebrahimi-etal-2018-hotflip}, a gradient-based method that generates adversarial examples on discrete text structure; PWWS \cite{ren-etal-2019-generating}, an importance-based method that substitutes words of high saliency. By contrast, black-box attacks assume no knowledge of the victim model’s architectures and parameters. Example techniques include the use of generative adversarial networks (GANs) \cite{zhao2018generating} and human-in-the-loop heuristics  \cite{wallace-etal-2019-trick}\\

\noindent{\textbf{Universal Triggers}: \citet{wallace-etal-2019-universal} generate universal attack triggers by using gradient signals to guide a search over the word embedding space. They are input-agnostic, which makes them more threatening in real-world scenarios. Despite being successful in confusing classification systems, universal triggers are often unnatural and can easily be detected by human readers. \citet{song-etal-2021-universal} generate attack triggers that appear closer to natural text by using a pre-trained GAN. Training a GAN in the ToS domain from scratch requires large datasets and GPU resources. In this work we try to generate natural triggers by simply skipping all the subword and special tokens during the search process; and leave the development and evaluation of a ToS-GAN to future work.}

\section{Universal Trigger Generation}
We assume a text input $x$ and its target label $y$ from the dataset $D = \{X, Y\}$, a trained victim classifier model $f$ that predicts $f(x) = \hat{y}$. While in a \textit{non-universal} targeted attack the focus is on flipping the prediction of a single text input $x$, our goal is to find an input-agnostic trigger $t$ consisting of a sequence of tokens $\{w_1, w_2,…,w_i\}$ such that when concatenating $t$ with any input $x$ from $X$, the victim model incorrectly predicts $f(x;t) = \tilde{y}$, where $\tilde{y} \neq \hat{y}$. Specifically, we use the following objective function: \\
\begin{equation}
    \argmin_{t} \EX_{x\sim X}[\mathcal{L}(\tilde{y}, f(x;t))]
    \label{equ:univ_loss}
\end{equation} 

To solve the above objective function, we follow the approach of \citet{wallace-etal-2019-universal}
by utilizing the HotFlip method \cite{ebrahimi-etal-2018-hotflip} at the token level: First, we initiate the trigger $t$ with a sequence of $i$ placeholder tokens (i.e., `the'); then we compute the gradient of (1) w.r.t the trigger. Since tokens are discrete, we approximate the loss function around the current token embedding using the first-order Taylor expansion
\begin{equation}
    \argmin_{e_i' \in \mathcal{V}} \left[ e_i' - e_{adv_i}\right]^T \nabla_{e_{adv_i}} \mathcal{L} 
    \label{equ:taylor}
\end{equation} 

\noindent{where $\mathcal{V}$ is the set of all token embeddings over the entire vocabulary and $e_{adv_i}$ represent the embedding of the current trigger token. 

We update the embedding for every trigger token $e_{adv_i}$ to minimize (\ref{equ:taylor}). 
This can be efficiently computed through  $d$-dimensional dot products, with $d$ corresponding to the dimension of the token embeddings. 
For constructing the entire updated trigger, we then use beam search to evaluate the top $i$ token candidates from (\ref{equ:taylor}) for each token position in the trigger $t$. 
As variable parameters, we run experiments with triggers of different lengths [3, 5, 8] and insert positions [begin, middle, end] in the input text. 

}

\section{Experiments}
\subsection{Dataset and the Victim Model}

The CLAUDETTE dataset \cite{lippi2019claudette, ruggeri2022detecting} consists of 100 ToS contracts (20,417 clauses) of online platforms. A clause is deemed as unfair if it creates an unacceptable imbalance in the parties’ rights and obligations, i.e., harms the user's rights or minimizes the online service's obligations. Each clause was labelled by legal experts. \footnote{To measure the inter-annotation agreement, \citet{lippi2019claudette} have an additional test set containing 10 contracts labelled by two distinct experts, which achieve a high inner-annotator agreement with standard Cohen $\kappa$= 0.871. For details on the annotation process and the legal rational of unfair contractual clauses, please refer the original CLAUDETTE paper. }

 Following \citealt{lippi2019claudette}, we discard sentences shorter than 5 words. In order to avoid an information leak between training and testing sentences by virtue of them stemming from the same document of contracts, we split the 100 contracts randomly into 40:40:20 for training, development and testing. Table \ref{tab:data_split} in Appendix \ref{app:data_split} shows the detailed statistics of each split. Notably, the CLAUDETTE has a very imbalanced class ratio of 9:1 (fair:unfair).

For the victim model, we finetune an instance of LEGAL-BERT (nlpaueb/legal-bert-base-uncased) \cite{chalkidis-etal-2020-legal} on the CLAUDETTE training set. Please refer to Appendix \ref{app:finetuning} for details on model finetuning. It achieves overall macro F1 of 88.9\%, 
97.7\% F1 for class fair, and 80.1\% for class unfair.

\subsection{Attack Results} 
In the following we focus on the attack scenario \textit{fairwashing}: targeted attacks that flip \textit{unfair} predictions to \textit{fair}. We apply the universal attack trigger algorithm on the development set and report the attack performance on the test set. The generated triggers can considerably degrade the victim model's performance. For instance, inserting the trigger of token length 8 ``\#\#purchased another opponent shall testify unless actuarial opponent'' in the middle of the sentence can decrease the model’s accuracy from 80.1\% to 16.9\%. However, we observe that triggers often contain special tokens or subwords, such as `[SEP]' or `\#\#purchased', which makes them easily detectable for human readers. Inspired by \citealp{wang2022global}, we facilitate the generation of natural triggers by simply skipping all subwords and special tokens during the search (hereafter we denote this approach as mode 'no\_subword' for simplicity). Although slightly less effective than the original triggers (Table \ref{tab:univ_trig_perf} in Appendix \ref{app:results}), the no\_subword triggers are less likely to be detected by human readers (See our human evaluation study in Section \ref{sec:humanEval}). 

\begin{figure}[]
    \centering
    \includegraphics[width =0.45\textwidth]{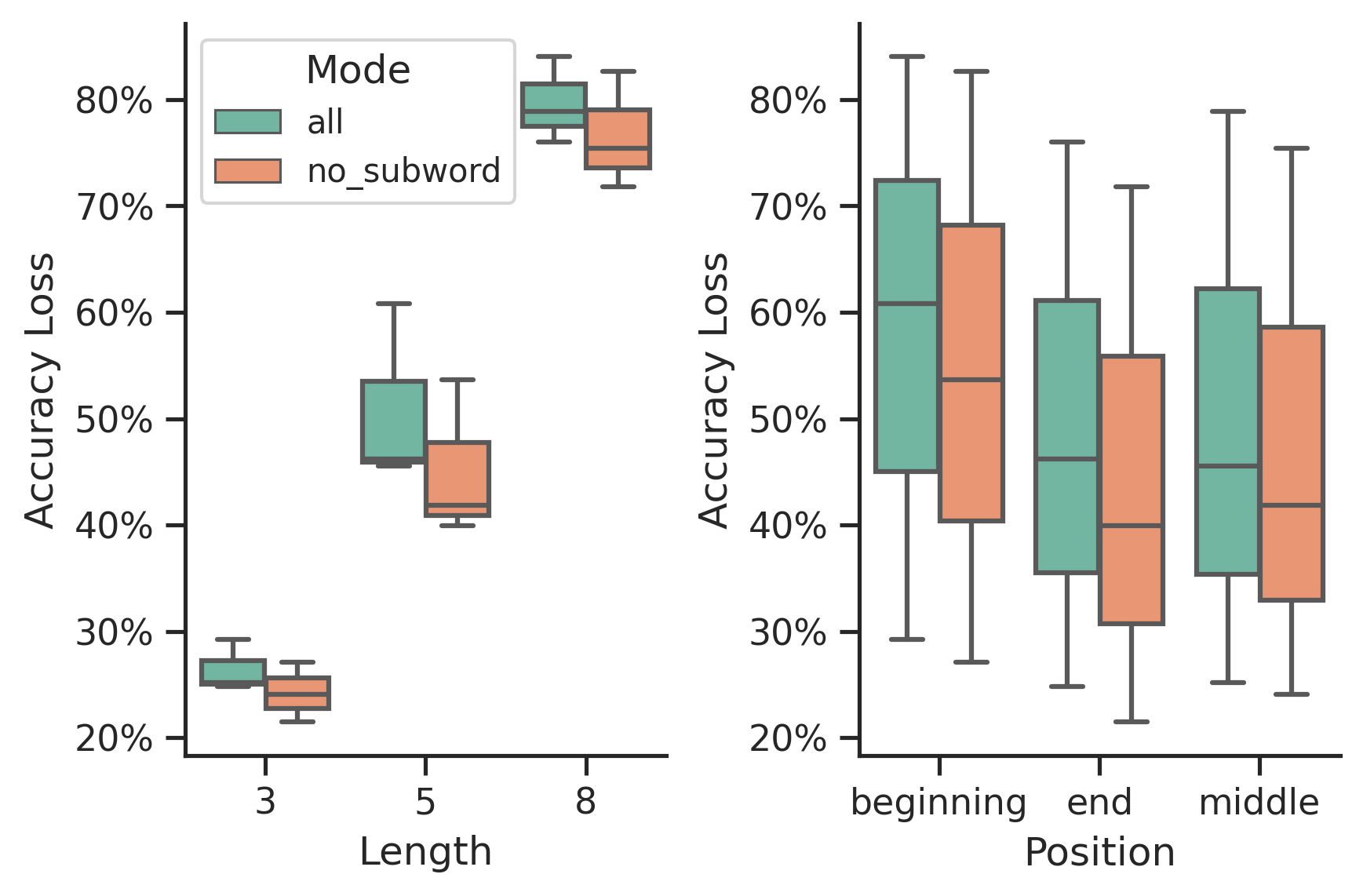}
    \caption{Accuracy loss of the victim model's detection performance when attacked by universal triggers of different insert positions and lengths. For completeness, we report the full attack results in Appendix \ref{app:results}}
    \label{fig:univ_attack_performance}
\end{figure}

We also run experiments to study the impact of trigger length, insert position, and mode (with/without subwords) on the attack's effectiveness. Figure \ref{fig:univ_attack_performance} shows that increasing the token length improves attack effectiveness by a noticeable margin. The victim model's accuracy degrades by ~25\% to ~60\% using three words and by ~80\% to ~13\% with eight words. The result also indicates the victim model's sensitivity to the insert position of the triggers. These results are consistent with previous studies \cite{wallace-etal-2019-trick, wang2022global}: Triggers are more effective when inserted at the beginning of the clause, which may be due to the transformer-based model paying more attention to the terms at the beginning of the text.
These results hold across both modes. Between the modes, a higher effectiveness is consistently observed for 'all' compared with 'no\textunderscore subword'. This is in line with 'no\textunderscore subword' generating triggers from a subset of potential trigger tokens of 'all' mode. 

\begin{figure*}[]
    \centering
    \includegraphics[width =\textwidth]{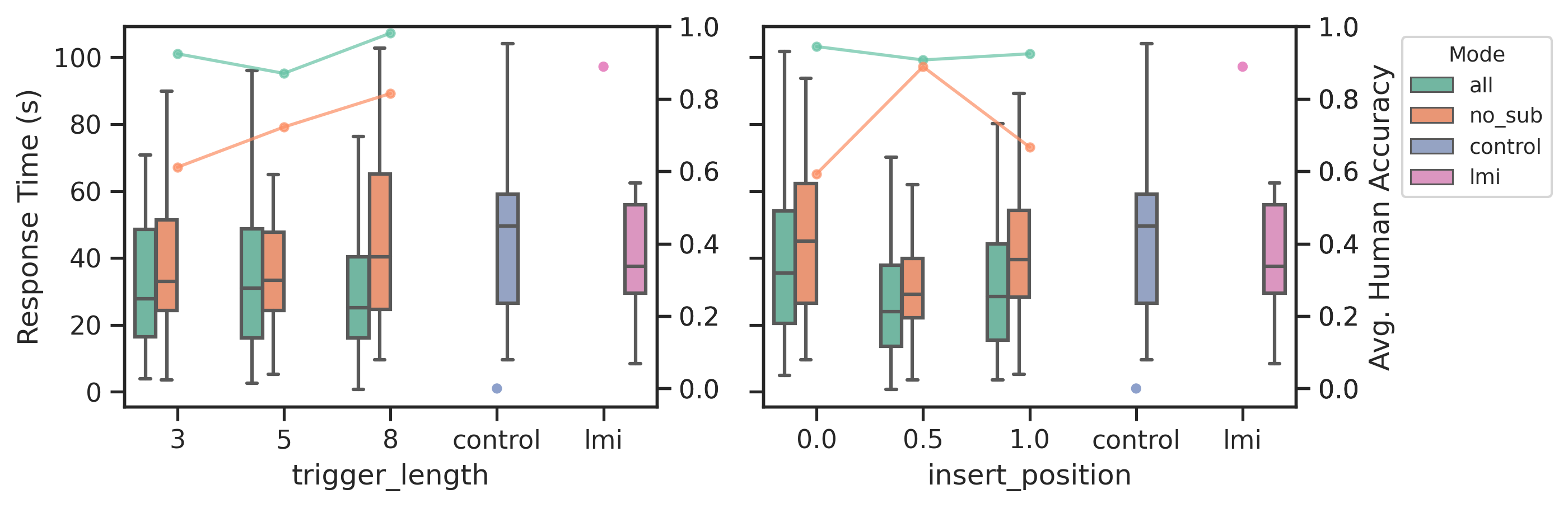}
    \caption{Human response time (box plots) and detection accuracy (line plots) for triggers of different insert positions and lengths.  \textit{Control} stands for the question where no trigger is inserted. \textit{LMI} represents an LMI trigger of length eight inserted in the middle of the sentence. The insert positions are the following. 0.0 : beginning, 0.5 : middle, 1.0 : end.}
    \label{fig:humanEval_response}
\end{figure*}

\section{Data Artifacts as Universal Triggers}
A growing number of works have raised awareness that deep neural models may exploit spurious artifacts in the dataset and take erroneous shortcuts \cite{mccoy-etal-2019-right,xu-2022}. In this section, we experiment with using dataset artifacts as universal triggers to explore the feasibility of generating universal triggers without access to the victim model’s gradient signals. 
Following \citet{gururangan-etal-2018-annotation}, \citet{wallace-etal-2019-universal} identified the dataset artifacts as words with high pointwise mutual information (PMI) \cite{church1990word} with each label.
Since the Claudette dataset has a heavily imbalanced label distribution, in order to prevent picking up very sparse tokens, in this work, we use local mutual information (LMI) \cite{schuster2019towards}, a re-weighted version of PMI. 
We observe that high LMI ranked words are successful triggers. We use the 8 highest LMI words and PMI words with label fairness as triggers (hereafter LMI trigger and PMI trigger respectively, please refer to Appendix \ref{app:lmi_triiger} for the list of words used); and insert them to the unfair clauses at different token positions. The LMI trigger is able to reduce the victim model’s classification accuracy from 80\% to around 60\%; while the PMI trigger can only reduce the performance to around 76\% (see Figure \ref{fig:lmi_attack} in Appendix \ref{app:lmi_triiger}). Although less successful than the universal adversarial triggers, the LMI triggers are natural and less detectable than 'all' mode triggers according to our human evaluation studies. Critically, LMI triggers are extracted by simply analyzing the training data and do not require access to the victim model. The attack effectiveness of LMI trigger highlights the potential threat of training data leakage in the NLP application.

\section{Human Evaluation Study}
\label{sec:humanEval}
We perform a human evaluation to study the impact of token length, insert position and mode on the triggers' detectability \footnote{We report the details of the web application used and full instructions for the human subjects in Appendix \ref{app:instruction}}. The task is to identify which sentence out of four candidate sentences from ToS contracts was modified. 
We include one question with no modified sentence as the control. 
In a previous study, \citealt{song-etal-2021-universal} directly asked the human participants to rate whether the generated triggers were natural or not. However, the rating of naturalness is very abstract and varies between individuals. Inspired by studies on the detection process in psychological studies \cite{pandya1995pattern,yap2007additive}, we assume response time (i.e., the length of time taken for a human to detect a trigger) can act as a proxy for the naturalness. To measure the human detectability of triggers, we hence collect the answer accuracy as well as the response time from the participants.\\
19 participants of different ages, English abilities, and legal experience were recruited from the personal network of the authors.   Figure (\ref{fig:humanEval_response}) demonstrates that it is consistently easier for participants to detect 'all' mode triggers than 'no\_subword' mode triggers. Participants were on average 19\% faster in detecting that a sentence inserted by ‘all’ than ‘no\_subword’ triggers; and they find 'all' triggers with 21\% higher accuracy on average. We include the LMI trigger of token length eight in the study and find its detectability is in between the ‘no\_subword’ and 'all' triggers of the same length. The intuitive notion that participants are better at finding longer triggers generally holds with regard to detection accuracy. Nevertheless, we cannot observe a trend in the response time change, which may be due to our small sample size.  Regarding the insert position, participants are the fastest in detecting triggers inserted in the middle. Further, we notice that special tokens and subwords make triggers more obvious. Qualitative, informal reports from participants indicate that 'spelling error' stuck out in a legal context. All triggers containing these tokens can be detected with more than 90\% accuracy, which include two 'all' triggers of length three (containing special token [SEP] or combination of subtokens '\#\#assignabilityconsult'); and one ‘no\_subword’ triggers inserted at position 5 (includes a bound stem ‘concul’). This likely explains why these two data points do not conform to the general trend of detection accuracy.

\section{Conclusion}
We attacked ToS unfair clause detectors with universal adversarial triggers generated by a gradient-based algorithm as well as by simply analyzing the training data. The effectiveness of the triggers exposes the vulnerability of the transformer-based classification model, and highlights the potential threat of training data leakage. We also conducted a human evaluation to study the detectability of the triggers. The results show that the triggers are less likely to be detected if they do not include subtokens. Future work can explore ways to generate more natural triggers in the legal domain, which may even deceive readers with a formal education in law.

\section*{Limitations}
\citet{wallace-etal-2019-universal} reduce the detection accuracy to 1\% while we can only manage to degrade it to 10\%. This might be due to the imbalanced label distribution and comparatively small size of the CLAUDETTE dataset. Our human evaluation is an initial exploration with only 19 participants. Future work will focus on using crowdsourcing techniques for large survey data collection.	Furthermore, we generate the 'no\_subword' triggers by skipping all the tokens preceded by the double hashtag '\#\#'. This enables us to avoid derivational morphemes and inflection suffixes but fails to exclude bound stems such as ‘consul’, which makes some triggers obvious to human readers. Future work can explore better ways to generate natural triggers. 

\section*{Ethics Statement}
The study presented here works exclusively with the publicly available CLAUDETTE dataset, which consists of the Terms of Service (ToS) Agreements of various online platforms. The techniques described in this paper are prone to misuse. However, we design this study to draw public attention to the vulnerability of the transformer-based classification model. We hope our work will help accelerate progress in detecting and defending adversarial attacks.
We finetuned the victim model and generated all the triggers on Google Colab. Our models adapted pretrained language models and we did not engage in any training of such large models from scratch. We did not track computation hours.

\bibliography{anthology}
\bibliographystyle{acl_natbib}

\newpage
\appendix
\section{Dataset Statistics}
\label{app:data_split}
Table \ref{tab:data_split} displays the statics of the CLAUDETTE dataset.

\begin{figure}[]
    \centering
    \includegraphics[width =0.5\textwidth]{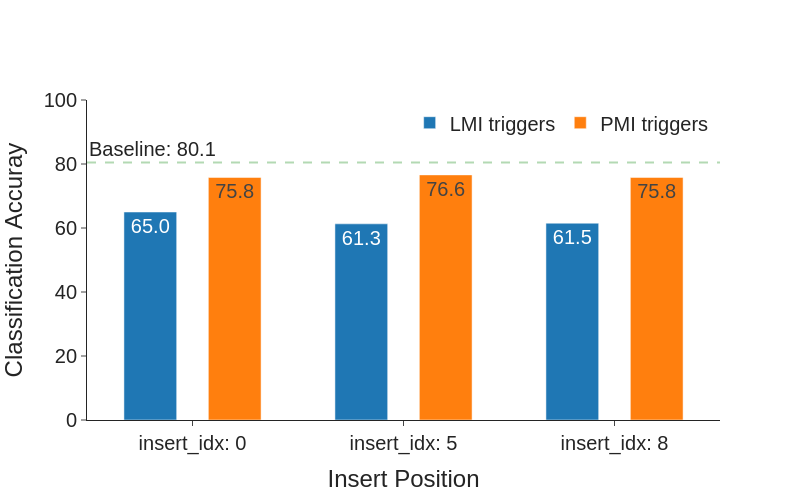}
    \caption{Attack performance of LMI trigger and PMI trigger of the different insert position.}
    \label{fig:lmi_attack}
\end{figure}

\section{Finetuning the Victim Model}
\label{app:finetuning}
We used LEGAL-BERT (nlpaueb/legal-bert-base-uncased) with a sequence classification head on top from the transformers library \cite{wolf2019huggingface}; and finetuned it on the CLAUDETTE  training set. The model is fine-tuned with 5 epochs, a learning rate of 1e-5. We determine the best learning rate using grid search on the development set and use early stopping based on the development set F1 score.

\begin{table}[]
\footnotesize
\begin{tabular}{|l|l|l|l|}
\hline
split & \# sentences & \% fair label & \% unfair label \\ \hline
train & 8354        & 89.5\%            & 10.5\%              \\ \hline
dev   & 8279        & 89.1\%            & 10.9\%              \\ \hline
test  & 3784        & 89.3\%            & 10.7\%              \\ \hline
\end{tabular}
    \caption{Statistics of the train, dev and test split of the CLAUDETTE dataset. }
    \label{tab:data_split}
\end{table}

\section{Additional Experimental Results}
\label{app:results}
Table \ref{tab:univ_trig_perf_fair} demonstrates the attack results on \textit{fair} clauses. Restricted to limited GPU resources, we generated only triggers of eight tokens which are inserted at the beginning of the sentence.

Table \ref{tab:univ_trig_perf} displays the attack on \textit{unfair} clauses with triggers of different lengths [3,5,8], insert position [begin, middle, end] and mode [original, no\_subword].

\begin{table*}[t]
\centering
\footnotesize
\begin{tabular}{|l|rrrrr|}
\hline
Trigger                                                                                                          & \multicolumn{1}{c}{Length} & \multicolumn{1}{c}{Position} & \multicolumn{1}{c}{Mode} & \multicolumn{1}{c}{Accuracy} & \multicolumn{1}{c|}{$\Delta$} \\
\hline
\hline
Baseline                                                                   & -                          & -             & - & 97.7                    &          \\
not liable whenever                   & 3                          & beginning                    &no\_sub                      & 69.8                        & -28.5\%                    \\
terminate our convening practices if     & 5                          & beginning                       &no\_sub                      & 47.6                        & -51.2\%                    \\
agree tankage bound through cloud terms 2016 laws & 8                          & beginning                          &no\_sub                      & 9.0                        & -90.0\%                    \\
\hline
\end{tabular}
\caption{Performance of Universal Triggers on Fair Clauses}
\label{tab:univ_trig_perf_fair}
\end{table*} 
\begin{table*}[t]
\centering
\footnotesize
\begin{tabular}{|l|rrrrr|}
\hline
Trigger                                                                                                          & \multicolumn{1}{c}{Length} & \multicolumn{1}{c}{Position} & \multicolumn{1}{c}{Mode} & \multicolumn{1}{c}{Accu.} & \multicolumn{1}{c|}{$\Delta$} \\
\hline
\hline
Baseline                                                                   & -                          & -                            & -                        & 80.1                        &                            \\
witness should testify                                                     & 3                          & beginning                    &no\_sub                    & 58.4                        & -27.0\%                    \\
may vote against                                                           & 3                          & middle                       &no\_sub                    & 60.8                        & -24.1\%                    \\
witness testified without                                                  & 3                          & end                          &no\_sub                    & 62.9                        & -21.5\%                    \\
interrelat order refusing priority where                                   & 5                          & beginning                    &no\_sub                    & 37.1                        & -53.7\%                    \\
consul must produce his attorney                                           & 5                          & middle                       &no\_sub                    & 46.6                        & -41.9\%                    \\
privilege to authenticate testimony groot                                  & 5                          & end                          &no\_sub                    & 48.1                        & -39.9\%                    \\
testimony allows contracts opposing person tuber testify where             & 8                          & beginning                    &no\_sub                    & 13.9                        & -82.7\%                    \\
compute another opponent shall testify unless lockbox opponent             & 8                          & middle                       &no\_sub                    & 19.7                        & -75.4\%                    \\
another witness seems thus admissible scope testify usc                    & 8                          & end                          &no\_sub                    & 22.6                        & -71.8\%                    \\
admissible in evidence                                                     & 3                          & beginning                    &all                      & 56.7                        & -29.3\%                    \\
\#\#assignabilityconsult assigned                                          & 3                          & middle                       &all                      & 59.9                        & -25.2\%                    \\
{[}SEP{]} expert testimony                                                 & 3                          & end                          &all                      & 60.2                        & -24.8\%                    \\
evid allowed equit testify where                                           & 5                          & beginning                    &all                      & 31.4                        & -67\%                    \\
{[}SEP{]} give precedence before priority                                  & 5                          & middle                       &all                      & 43.6                        & -45.6\%                    \\
368 hearsay witnesses may exclude                                          & 5                          & end                          &all                      & 43.1                        & -46.2\%                    \\
inference forbid 2028 opposing person may testify where                    & 8                          & beginning                    &all                      & 12.8                        & -84.0\%                    \\
\#\#purchased another opponent shall testify unless actuarial opponent     & 8                          & middle                       &all                      & 16.9                        & -78.9\%                    \\
assist {[}SEP{]} witness normally justifies cross admissibilitywillingness & 8                          & end                          &all                      & 19.2                        & -76.0\%                    \\
\hline
\end{tabular}
\caption{Performance of Universal Triggers on Unfair Label}
\label{tab:univ_trig_perf}
\end{table*}

\section{Instruction for the human evaluation study}
\label{app:instruction}
\begin{figure*}[]
    \centering
    \includegraphics[width =\textwidth]{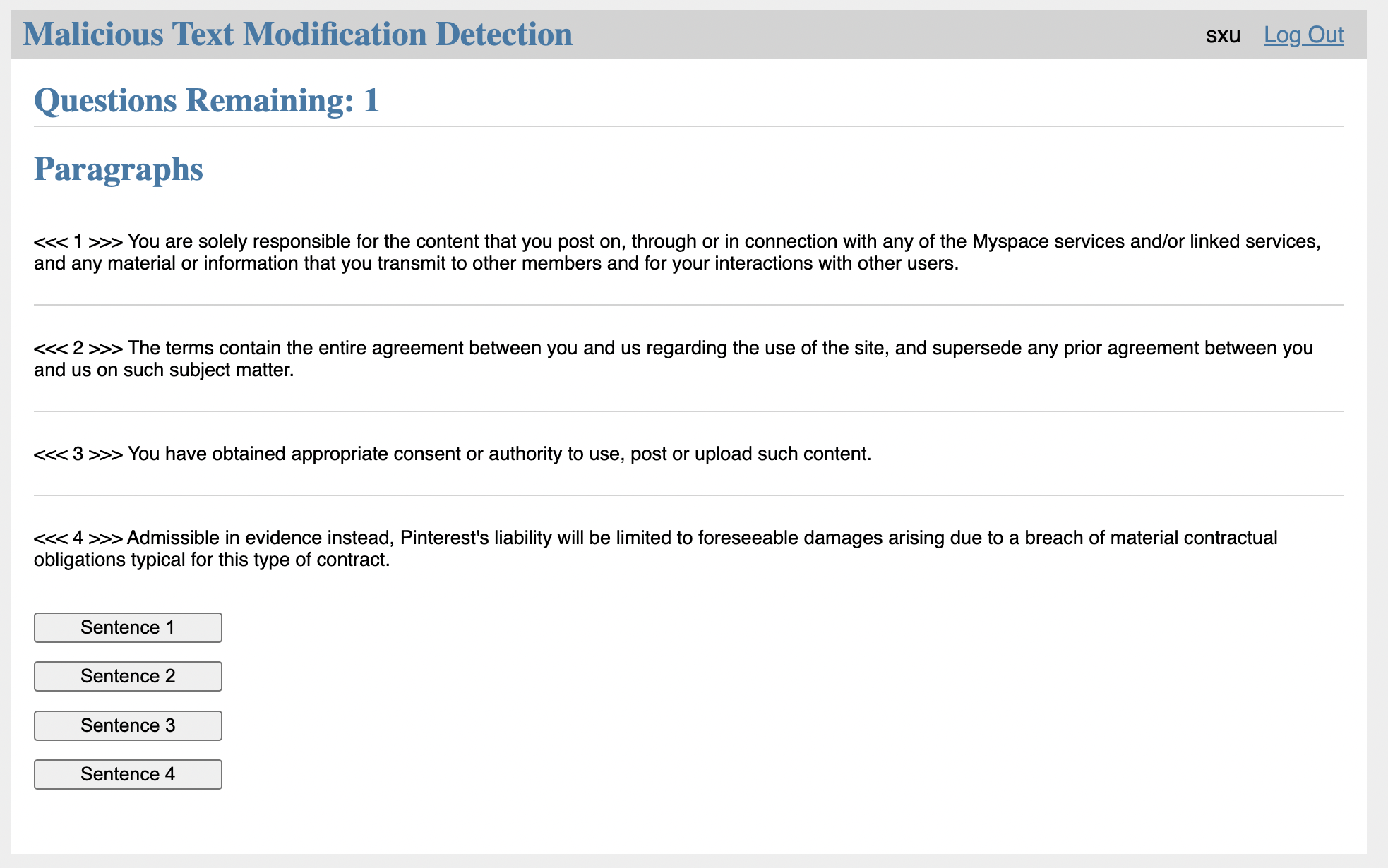}
    \caption{Screenshot of the web application for human evaluation}
    \label{fig:website}
\end{figure*}

The application is written in Python using Flask \cite{grinberg2018flask} and was hosted on an AWS EC2 instance. It included a landing page with a short instructions. Figure \ref{fig:website} is a screenshot of the web application. Following is the instruction on the landing page for the human evaluation study:

\noindent{\textbf{``Background information}\\
When using online platforms, users are asked to agree to the Terms of Service (ToS). ToS documents tend to be long and difficult to understand. As a result, most users accept the terms without reading them, including clauses which would be deemed unfair under consumer protection standards.
Therefore, applications that can support consumers in detecting unfair clauses would be useful. Nevertheless, studies have shown that such applications are vulnerable to adversarial attacks; even slight modifications to the input text, like inserting a few words into the text, can cause incorrect classifications. In this study, we ask you to help us detect the malicious modifications in the text.\\}
\textbf{Task instruction}\\
You will be shown an excerpt of four sentences from a ToS contract. The task is to identify which sentence is modified. Please feel free to contact us if you have any questions. Many thanks for taking part in the study.''

\section{LMI and PMI triggers}
\label{app:lmi_triiger}
Figure \ref{fig:lmi_attack} demonstrates the attack performance of LMI and PMI triggers. 
The 8 highest LMI ranked words that used as LMI trigger are ['information', 'payment', 'must', 'provide', 'person', 'license', 'rights', 'please']. The PMI trigger words are: ['berlin', 'attribution', 'addressing', 'android', 'sources', 'organiser', 'pc', 'unreasonable']  \\

\end{document}